\documentclass{article}
\usepackage[preprint]{neurips_2025}

\usepackage[toc,page,header]{appendix}
\usepackage{minitoc}

\usepackage[utf8]{inputenc} 
\usepackage[T1]{fontenc}    
\usepackage{hyperref}       
\usepackage{url}            
\usepackage{booktabs}       
\usepackage{amsfonts}       
\usepackage{nicefrac}       
\usepackage{microtype}      
\usepackage{xcolor}         
\usepackage{microtype}
\usepackage{graphicx}
\usepackage{subfigure}
\usepackage{booktabs} 
\usepackage[most]{tcolorbox}

\usepackage{enumitem}
\usepackage{amssymb}
\usepackage{url}
\usepackage{pifont}

\title{What's the next frontier for Data-centric AI? \\ Data Savvy Agents!
}

\author{%
  Nabeel Seedat\thanks{Equal contribution.} \\
  University of Cambridge \\
  \texttt{ns741@cam.ac.uk} \\
  \And
  Jiashuo Liu\textsuperscript{*} \\
  Tsinghua University \\
  \texttt{liujiashuo77@gmail.com} \\
  \And
  Mihaela van der Schaar \\
  University of Cambridge \\
  \texttt{mv472@cam.ac.uk} 
}

\begin{document}

\doparttoc
\faketableofcontents

\maketitle

\begin{abstract}
The recent surge in AI agents that autonomously communicate, collaborate with humans and use diverse tools has unlocked promising opportunities in various real-world settings. However, a vital aspect remains underexplored: how agents handle data. Scalable autonomy demands agents that continuously acquire, process, and evolve their data.
In this paper, we argue that data-savvy capabilities should be a top priority in the design of agentic systems to ensure reliable real-world deployment. Specifically, we propose four key capabilities to realize this vision:
(1) \emph{Proactive data acquisition}: enabling agents to autonomously gather task-critical knowledge or solicit human input to address data gaps;
(2) \emph{Sophisticated data processing}: requiring context-aware and flexible handling of diverse data challenges and inputs;
(3) \emph{Interactive test data synthesis}: shifting from static benchmarks to dynamically generated interactive test data for agent evaluation; 
and (4) \emph{Continual adaptation}: empowering agents to iteratively refine their data and background knowledge to adapt to shifting environments.
While current agent research predominantly emphasizes reasoning, we hope to inspire a reflection on the role of data-savvy agents as the next frontier in data-centric AI.
\end{abstract}

\section{Introduction}

Imagine a world where AI systems seamlessly collaborate with humans, autonomously gathering and analyzing vast amounts of data, continuously adapting to shifting environments, and providing real-time insights to inform critical decisions. 
Such agents could aid domains from healthcare to climate policy, driving smarter decisions, tailored education, and faster scientific discoveries. They would act as trusted partners, augmenting human decision-making. 

This vision of AI, capable of self-learning and autonomous action, is no longer purely speculative. 
Recent developments in Large Language Models (LLM)-based agents have brought us closer to this future, with these systems demonstrating impressive abilities in natural language understanding, problem solving and even tool use. However, despite their remarkable progress, current LLM agents remain constrained. 
They mainly operate in controlled environments with predefined data and well-structured tasks, and they rely heavily on static datasets and benchmarks. 
However, today's AI agents are increasingly expected to operate in open-ended, dynamic environments—whether for scientific discovery, industrial automation, finance, or healthcare.
In these scenarios, data presents multiple challenges: it is often \emph{incomplete}, requiring proactive information seeking; \emph{messy and noisy}, demanding sophisticated diagnostic and processing capabilities; \emph{constantly evolving}, necessitating continuous knowledge updates; and \emph{difficult to evaluate} through traditional static benchmarks.
This leaves a critical gap: real-world scenarios demand more than predefined knowledge and tasks—they require systems that can actively engage with dynamic, noisy, and evolving data in real-time.

\begin{figure*}[t]
    \vspace{-10mm}
	\centering\includegraphics[width=0.75\textwidth]{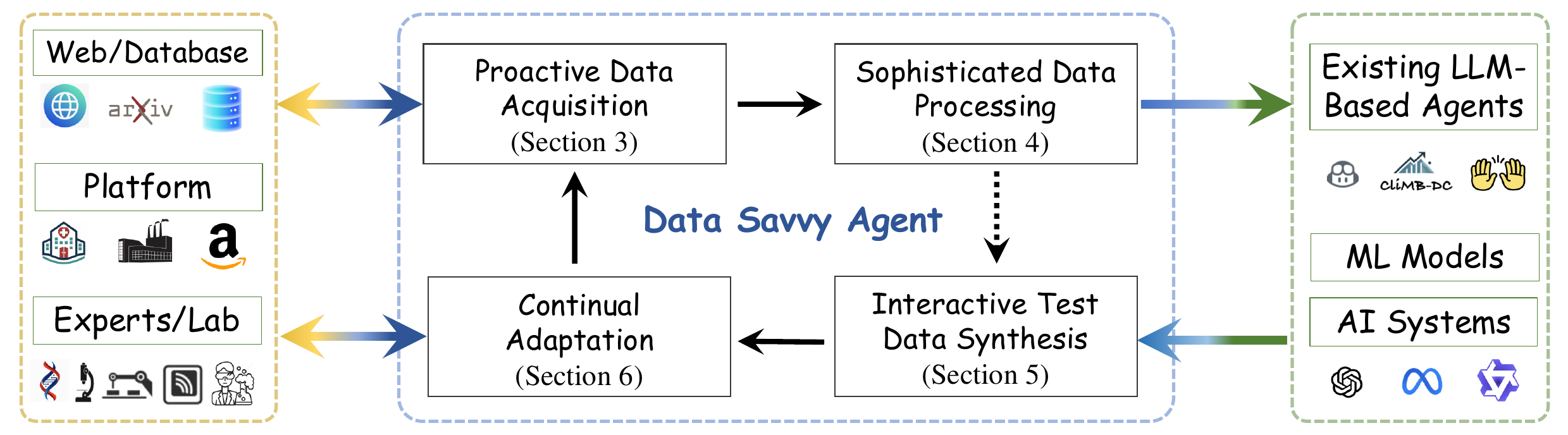}
	\vspace{-0.1in}
    	\caption{\small Framework of a data-savvy agent, built on four key capabilities: \emph{proactive data acquisition} (Sec.~\ref{sec:data-acquisition}), \emph{sophisticated data processing} (Sec.~\ref{sec:data-processing}), \emph{interactive test data synthesis} (Sec.~\ref{sec:data-evaluation}) \& \emph{continual adaptation} (Sec.~\ref{sec:adaptation})}
    \label{fig:overview}
  \vspace{-0.1in}
     \rule{\linewidth}{.5pt}
         \vspace{-10mm}
\end{figure*}

To bridge this gap, we propose a new research direction: the development of data-savvy agents, building on the common ground of agent-based AI (focused on decision-making and automation) and data-centric machine learning (focused on dataset curation in static settings).
These data-savvy agents would go beyond merely processing data—they would autonomously acquire, refine, and adapt their knowledge. 
Advancing autonomous data management would let data-savvy agents thrive in complex settings, becoming flexible, resilient, and self-improving.

\textbf{\emph{In this paper, we argue that data-savvy capabilities should be a priority in the design of future agentic systems}}. 
Rather than merely executing tasks with pre-existing data, we highlight four key areas that need focused research: \emph{proactive data acquisition}, \emph{sophisticated data processing}, \emph{interactive test data synthesis}, and \emph{continual adaptiveness}.
This paradigm shift unlocks exciting opportunities across multiple research communities: \textbf{{\textcircled{1}}} \textbf{Agent researchers} can extend their focus beyond reasoning and tool use to include dynamic data acquisition, adaptive learning, and real-time knowledge integration, making agents more autonomous and resilient in complex environments; \textbf{{\textcircled{2}}} \textbf{Data-centric ML researchers} can explore new data challenges in open-ended, interactive, and non-stationary environments; \textbf{{\textcircled{3}}} \textbf{Broader scientific researchers} and \textbf{industrial engineers} can leverage data-savvy agents to accelerate discoveries, optimize large-scale industrial systems, and develop AI-driven solutions that continuously learn and evolve alongside real-world complexities.

We highlight that, as a paper, our goal is to spark discussion on data-savviness as a foundational capability for AI agents, rather than providing explicit solutions.  Moreover, given the broad potential of data-savvy agents, prescribing a single architecture isn't appropriate nor feasible as optimal design is context dependent based on: (i) Computational constraints (mobile vs. cloud systems), (ii) Data requirements (real-time vs. batch), (iii) Cost and resources, (iv) Application-specific needs. 

Consequently, we hope this paper stirs debate within the ML research community, challenging the idea that the future of AI will be defined not just by what agents can do—but by what data they can understand, shape, and evolve with --- as the path toward truly autonomous agents.
\vspace{-3mm}
\section{Data-Savvy Agent}
\vspace{-2mm}
We begin by providing an overview of the role of data-savvy agents. As illustrated in Figure~\ref{fig:overview}, data-savvy agents fill the vacancy between real-world data sources and general LLM-based agents or traditional ML models. Acting as a crucial bridge, they enable the seamless integration of diverse data streams into AI systems. On one side, data-savvy agents interact with a wide range of data sources, from easily accessible public web databases and open repositories to platform-based data that requires specialized devices or infrastructure, such as hospital records, industrial systems or smart devices. 
They also manage scarce, high-value experimental data from labs and proprietary sources.
On the other hand, data-savvy agents feed curated real-time data to AI systems and run interactive auto-evaluations that facilitate adaptation.

To enable this, we posit that data-savvy agents must be equipped with four key capabilities. First, they need \emph{proactive data acquisition}, which involves handling raw, messy, and dynamic data from various sources to gather application-specific data or knowledge. Second, they require \emph{sophisticated data processing}, enabling them to manage and curate diverse data types in a context-aware manner. Third, they must have the ability to conduct \emph{interactive test data synthesis} as well as auto-evaluation, where dynamically generated interaction data helps reliably evaluate agent performance. Finally, data-savvy agents should be capable of \emph{continual adaptation}, which entails iteratively refining data or acquisition strategies to adjust to shifting environments, thus enhancing model performance over time. We detail these capabilities next. Additionally, we provide a structured breakdown of the key research directions and their levels of difficulty in Appendix \ref{appendix:actionable}.

By integrating these capabilities, data-savvy agents go beyond data-centric AI’s focus on improving static datasets or well-structured tasks, to proactive and adaptive data engagement, enabling AI agents to operate in dynamic, real-world settings. For more details on this contrast, see Appendix \ref{appendix:comparison}.

\section{Capability 1: Proactive Data Acquisition}
\label{sec:data-acquisition}
\vspace{-2mm}
\begin{tcolorbox}[
    colback=blue!2!white,    
    colframe=blue!50!black,  
     breakable,
    fonttitle=\bfseries,
    boxrule=.5pt,
    left=5pt,            
    right=5pt, 
    top=1mm, bottom=1mm,
    arc=1mm
]
\emph{Proactive data acquisition} denotes an agent's autonomous ability to collect relevant data via:
\begin{itemize}[nosep,left=0pt]
    \item Automated data acquisition from publicly accessible web sources;
    \item Targeted data extraction \& aggregation from diverse unstructured databases;
    \item Strategic interactions \& experimentation with domain experts and platforms.
\end{itemize}
\end{tcolorbox}
\vspace{-0mm}
This capability emphasizes \emph{resource-intensive}, \emph{dynamic}, and \emph{strategically planned} data gathering, distinguishing it from the relatively straightforward and static nature of information retrieval, as seen in retrieval-augmented generation (RAG) approaches~\citep{gao2023retrieval}. 
Unlike retrieval, which operates on predefined and accessible knowledge bases, data acquisition involves navigating unstructured or restricted data sources, adapting to evolving data landscapes, and managing significant logistical and financial constraints.
For a more detailed comparison, please refer to Section~\ref{subsec:data-acquisition-limitations}.

\subsection{Why It Matters?}
\label{subsec:data-acquisition-why}
Quality data underpins AI; poor data undermines even advanced models. For example, ImageNet~\citep{deng2009imagenet} transformed deep learning with large-scale labeled images, while industries like recommendation systems, LLM pretraining, and advertising invest heavily in data collection.

In niche fields such as industrial applications, chemistry, and material science, data acquisition presents significant challenges compared to easily accessible web data. Accessing such data often requires application-specific knowledge, specialized facilities, or even human expertise and experimentation. In fact, for most real-world applications, collecting suitable data is frequently the greatest challenge, often outweighing the complexity of algorithm or model development.
Let's consider examples across two domains to demonstrate this: \emph{Environmental monitoring} in pollution source tracking exemplifies fragmentation challenges. Models must reconcile satellite imagery with ground sensor networks, while navigating incompatible agency reporting formats (PDFs vs. APIs). The rarity of labeled crisis data, like chemical spills, further compounds the difficulty of training.
Alternatively, in the case of \emph{industrial diagnostics} to predict equipment failure, there can be coverage and quality issues. Legacy sites have few sensors and often suffer from drift and bad logs. Safety protocols and production demands create additional blind spots by restricting when and where data can be collected.

In summary, these hurdles highlight that real-world data collection is rarely straightforward and is frequently shaped by the unique complexities of the application, making it an inherently difficult and nuanced task.
As a result, there is a strong practical demand for autonomous data acquisition.

\subsection{Current Progress \& Limitations}
\label{subsec:data-acquisition-limitations}
Given the critical importance of data acquisition, we begin by reviewing the latest advancements in ``traditional data acquisition'' within the data-centric AI community.
We then highlight the gap between current LLM-based agents and the goal of proactive data acquisition, emphasizing how this crucial aspect has been overlooked and why addressing it is essential—even for the agents themselves.\\\\
\textbf{Traditional Data Acquisition via Active Learning.}
Existing data acquisition primarily focuses on \emph{simulated} or \emph{idealized} settings, where data is often assumed to be pre-collected or available in well-structured and standardized formats. 
For instance, \emph{active learning}~\citep{settles2009active, settles2011theories,konyushkova2017learning} addresses the problem of iteratively selecting data points from a large (typically unlabeled) data pool for labeling \citep{astorga2024active}.
However, this approach assumes full access to the entire unlabeled data pool, an assumption that does not hold in practice, as the full dataset is typically not visible to the data acquirer.
While recent works~\citep{chen2023data,hedomain} on data acquisition relax the need for full access, they still assume the existence of multiple data sources or providers and primarily focus on efficiently selecting among them. 
However, this is a secondary problem compared to real-world data acquisition, where \emph{the most significant challenge lies in acquiring a comprehensive data pool} in the first place—as illustrated in Section~\ref{subsec:data-acquisition-why}.

In addition to traditional data acquisition, recent LLM-based agents have explored information retrieval---a related but simpler, preliminary step in proactive data acquisition.\\\\
\textbf{Retrieval-Augmented Generation for LLMs}
Recently, retrieval-augmented generation (RAG) for LLMs has seen rapid development. This approach enhances LLMs by retrieving relevant document chunks from external knowledge bases during inference, fine-tuning, and even pretraining. The knowledge base typically consists of structured databases or well-curated sources like Wikipedia and arXiv~\citep[Table 1]{gao2023retrieval}, enabling the LLM agent to retrieve the most relevant documents for more accurate and contextually grounded generation.
However, data acquisition is \emph{far more complex} than information retrieval, particularly in the following aspects:
\begin{itemize}[nosep]
	\item Data Accessibility: Unlike retrieval, where the knowledge base is predefined and easily accessible, data acquisition often involves navigating complex website structures, dynamic data sources, and negotiating access to proprietary or restricted datasets. Simple search is often insufficient, requiring more sophisticated techniques to extract relevant data.
	\item Data Quality and structure: Retrieval assumes well-structured and high-quality data, whereas acquisition deals with raw, unstructured, or incomplete data, requiring preprocessing.
	\item Dynamic and Evolving Sources: Retrieval operates on static and pre-defined knowledge bases, while acquisition must account for real-time, dynamic data sources.
	\item Cost and Scalability: Acquiring large-scale datasets often incurs substantial costs and logistical challenges, unlike retrieval, which leverages existing repositories.
\end{itemize}
These challenges highlight the limitations of traditional active learning, which often overlooks these practical issues. Additionally, we argue that RAG for LLMs also necessitates proactive data acquisition to enrich and broaden the knowledge base—a point we explore in detail in Section~\ref{subsec:impact-agent}.

\subsection{Research Directions}
\label{subsec:data-acquisition-direction}
Based on the type of data acquisition—such as whether the data can be obtained from the web or relies on instruments, equipment, or human effort—we outline research directions to empower data-savvy agents on proactive data acquisition.

\textbf{Direction 1: Application-Specific \emph{Web Data Acquisition}.}\quad
For domains such as biology, chemistry, and climate science, data is often messy, exists in diverse formats, and originates from soures with complex structures, highlighting the need for application-specific data acquisition.
We emphasize the following key capabilities that must be addressed:
\begin{itemize}[nosep]
	\item[1.] \textbf{Application-Specific Data Source Discovery}: The ability to autonomously identify and explore relevant data sources from the web, especially those with \emph{application-specific structures} or \emph{domain-specific nuances}. 
	Many specialized websites, such as biological databases or scientific repositories, have unique architectures, including dynamic content, nested navigation, or specialized query interfaces. 
	A data-savvy agent must interpret these structures, such as handling advanced web elements (e.g., form submissions, API calls) and filtering sources based on relevance and accessibility. 
Additionally, the discovery process should integrate domain-specific rules and human feedback to identify valuable sources.
	\item[2.] \textbf{Structured Data Extraction}: The capability to extract data from complex and heterogeneous formats, such as HTML tables, JSON files, or PDF documents. This requires advanced parsing techniques, optical character recognition (OCR) for scanned documents, and machine learning models to handle diverse layouts. 
	\item[3.]  \textbf{Adaptive Acquisition Strategy}: The ability to dynamically adjust acquisition strategies based on real-time feedback or constraints encountered during the process. For example, the agent should recognize when a data source becomes inaccessible, requires alternative access methods (e.g., API authentication), or needs additional domain-specific context to proceed. This ensures the agent can maintain robust performance across varying conditions and prioritize high-value data sources while minimizing time and resource consumption.
\end{itemize}

\textbf{Direction 2: \emph{Experimental Data Acquisition} via Human or Platform Interaction.}\quad 
When dealing with experimental data in scientific domains like biology, chemistry or materials, it is often not possible to obtain the required data from the web. 
Instead, agents must interact with human scientists or the platforms for data collection, which could involve the agent assisting in designing experiments and measurement strategies. 
By engaging in iterative and context-aware dialogue with human experts, the agent can bridge the gap between computational models and experimental practices, ensuring alignment with the specific requirements of the task at hand.
This requires the agent being able to \emph{understand domain-specific experimental constraints} and \emph{adapt their guidance based on feedback or evolving experimental outcomes}. 
Through such human-agent collaboration, experimental data acquisition can become more efficient, targeted, and aligned with real-world challenges.

\section{Capability 2: Sophisticated data processing}
\label{sec:data-processing}
\vspace{-2mm}
\begin{tcolorbox}[
    colback=blue!2!white,    
    colframe=blue!50!black,  
     breakable,
    fonttitle=\bfseries,
    boxrule=.5pt,
    left=5pt,            
    right=5pt, 
    top=1mm, bottom=1mm,
    arc=1mm
]
\emph{Sophisticated data processing} denotes an agent's autonomous ability to handle complex, real-world data using the following capabilities:
\begin{itemize}[nosep, left=0pt]
    \item Diagnose and resolve data issues.
    \item Adapt processing to contextual nuances and domain specific requirements.
    \item Integrate and appropriately use advanced data-centric tools.
    \item Reason both autonomously and with humans about data quality challenges.
\end{itemize}
\end{tcolorbox}
\vspace{-0mm}
This capability emphasizes an agent’s ability to reason about and handle complex, real-world data beyond standard preprocessing pipelines. 
\subsection{Why It Matters?}
The promise of autonomous AI agents in tasks from drug discovery to market analysis can only be fulfilled insofar as their ability to handle the messy, dynamic reality of real-world data \citep{zha2023data, seedat2024dips, seedat2023triage,kumar2024opportunities,seedat2023navigating} --- realizing this requires the capability for \emph{sophisticated data processing}. Specifically, real-world data is often noisy, biased, ambiguous, context-dependent and constantly shifting
\citep{seedat2023navigating,renggli2021data,aroyo2022data,sambasivan2021everyone,jain2020overview, seedat2024matchmaker}. Consider the challenge in finance, where an agent analyzing stock market data, to guide actions, must distinguish between \emph{missing values} caused by technical glitches versus deliberate trading halts - a nuance lost on static data processing pipelines that treat all gaps as noise.
However, the need for data savviness is further complicated in cases of data issues that might exist in open-ended environments like the web. e.g. an agent trying to crawl for information about a restaurant would need nuanced capabilities to handle complex real-world data, which might include AI-generated spam, biased user reviews, or outdated information \citep{verge_ftc_fake_review_2024,read_drowning_2024}. This demands reasoning about what is signal and noise.

\subsection{Current Progress \& Limitations.}
\label{subsec:processing-limitations}
\textbf{Current Data-Centric Tools.}\quad
The data-centric ML community has introduced various tools to tackle common data issues, including imputation \citep{jarrett2022hyperimpute}, data cleaning \citep{northcutt2021pervasive} and data valuation tools \citep{seedatdissecting,jiang2023opendataval, seedat2023triage, seedat2024dips, seedat2022diq}.
However, real-world data issues are far more complex for two key reasons: (1) as noted earlier, many data problems are context-dependent and require domain-specific knowledge; and (2) these issues often co-occur and are inter-correlated, meaning multiple tools must be integrated to effectively address complex scenarios.
Moreover, despite such tools, there is a significant gap in terms of integrated, automated pipelines capable of deploying in specialized domains.

We review two types of AI agents and how they currently handle data: (1) those that process data for modeling and predictive purposes; (2) those that handle data from open-ended tasks.

\textbf{Agents processing data for modeling.} Here we consider agent based systems like Data Interpreter \citep{hong2024data}, DS-Agent \citep{guo2024dsagent}, CleanAgent \citep{qi2024cleanagent}, GPT Code Interpreter, CliMB \citep{saveliev2024climb} etc. The primary challenge is that these agents often default to rigid pipelines based on standard data science practices. While appropriate for standard problems, in more complex cases, simply applying standard pipelines without reasoning about the data and context might lead to failures. 
Consider the following failure example: a healthcare agent imputes missing blood pressure values using population averages, unaware that missingness correlates with patient severity. As shown in \cite{van1999multiple}, doing so would result in a model underestimating mortality risk and affect the outcomes and decisions based on the survival analysis.

These limitations stem from agents treating data processing as a \textit{procedural checklist} rather than \textit{contextual reasoning}. Moreover, current agents optimize for workflow completion (e.g. Data interpreter is based on code execution success) over understanding domain-specific challenges.  Finally, despite immense progress in tooling and method development from the data-centric ML research community --- current agents often do not integrate said state-of-the-art tools.

\textbf{Agents processing data for open-ended cases.} AI agents might also be used for open-ended tasks in contrast to their use for data science, software engineering and ML pipelines. Let's consider the case of web agents designed to autonomously navigate the web, interact with websites, and process data to complete tasks such as information retrieval, data extraction, and task automation \citep{he2024webvoyager,koh2024visualwebarena}. 
However, a data processing challenge for agents is that web content is often unstructured, dynamic, and noisy. Hence, agents must process visual elements along with messy HTML and JavaScript. Additionally, while agents focus on automation, a neglected aspect in open-ended tasks like web surfing is that agents must distinguish between signal and noise, such as filtering out irrelevant ads, AI-generated spam, or outdated information. This requires sophisticated data processing capabilities and context-aware reasoning.

\subsection{Research Directions}  

We highlight the following research directions for the ML community necessary to endow data savvy AI agents with the capabilities of sophisticated data processing.

\textbf{Direction 1:} Reimagining agent data processing as \emph{dynamic and context-aware}.

\begin{enumerate}[nosep]
    \item \textbf{Diagnose Interdependent Issues}: Agents should be able to detect and resolve composite challenges like missingness combined with temporal leakage. This would require improvements to agent's causal reasoning \cite{xiong2024improving}, and context-aware diagnosis \citep{du2024survey, sarker2022context, dey2018context}.
    
    \item \textbf{Orchestrate Adaptively}: Agents should be able to sequence actions or tool usage in an adaptive and context-dependent manner, accounting for domain constraints. Advancements in task decomposition are vital for this \citep{gabriel2024advancing,rasal2024navigating}. 
    
\end{enumerate}

\textbf{Direction 2:} Reimagining human-agent collaboration for data processing with human experts.

\begin{enumerate}[nosep]
    \item \textbf{Expert translation}: Agents should have the capability to translate human requirements into executable and verifiable rules via natural language interaction.
    
    \item \textbf{Expert alignment}: Agents should have the capability to align with domain experts. In particular, agents should both ascertain when is the opportune time to prompt experts for information/feedback \citep{feng2024large}, as well as, have the capability to correct based on this feedback. For instance, in finance, traders can recognize valid "Black Swan" market anomalies (e.g., flash crashes) such that an agent does not misclassify it as an outlier to remove.
\end{enumerate}

\textbf{Direction 3:} Integration of data-centric ML research tools. Current AI agents often default to basic tooling when processing data. Hence, it is vital that future data savvy AI agents incorporate tooling advancements from the data-centric ML research community. 

\section{Capability 3: Interactive Test Data Synthesis}
\label{sec:data-evaluation}

\vspace{-2mm}
\begin{tcolorbox}[
colback=blue!2!white,
colframe=blue!50!black,
breakable,
fonttitle=\bfseries,
boxrule=.5pt,
left=5pt,
right=5pt,
top=1mm, bottom=1mm,
arc=1mm
]
\emph{Interactive Test Data Synthesis} refers to an agent’s ability to autonomously generate, refine, and manage evaluation data tailored to specific tasks and domains. This process includes:
\begin{itemize}[nosep,left=0pt]
\item Context-aware generation of synthetic test data tailored to specific applications or tasks;
\item Adaptive integration of feedback loops from domain experts to refine and improve data relevance and evaluation accuracy;
\item Generation of interactive, human-like test cases for testing scenarios that require nuanced communication or multi-turn dialogues.
\end{itemize}
\end{tcolorbox}
\vspace{-0mm}
This capability emphasizes the importance of \emph{data-centric} test generation, blending human insights and synthetic data to continuously refine evaluation to align with real-world applications.

\subsection{Why It Matters?}
Effective evaluation is vital to enhancing AI system capabilities. 
Without proper test data, diagnosing weaknesses and identifying opportunities for improvement becomes nearly impossible. 
However, real-world evaluation is far from simple with the following challenges:
\vspace{-0.05in}
\begin{itemize}[nosep]
\item[1.] \textbf{Scarcity of High-Quality Test Data}: Unlike traditional tasks with readily available data (e.g., image classification), real-world applications often face limited, fragmented, and noisy test data. This scarcity makes generating reliable evaluation data both critical and challenging  (as mentioned in Section~\ref{subsec:data-acquisition-why}).
\item[2.] \textbf{Complexity of Tasks and Domains}: Agentic systems, being designed for broad use cases, must be evaluated across a wide spectrum of tasks. For instance, evaluations might span diverse domains, such as software engineering, healthcare, and customer service, which require domain-specific test cases and scenarios~\citep{liuagentbench,xu2024theagentcompany}.
\item[3.] \textbf{Human-in-the-Loop Evaluation}: Modern agentic systems increasingly require human collaboration for evaluation. However, involving humans in testing introduces scalability issues and demands real-time interaction, which complicates large-scale, automated evaluation.
\end{itemize}
\vspace{-0.05in}
These challenges highlight the importance of \emph{automated and adaptive test data synthesis}. The ability to dynamically create and refine test cases ensures that evaluations are both efficient and representative of real-world applications, empowering systems to adapt and improve faster.

\subsection{Current Progress \& Limitations}
In addition to the lack of real-world test data (see Section~\ref{subsec:data-acquisition-limitations}), we highlight recent challenges in agentic system evaluation, especially data generation for complex, human-involved tasks.

\textbf{Manual Task Generation for Evaluation.}\quad
Existing benchmarks for agentic systems, including tasks from diverse fields such as software engineering and gaming, are curated manually~\citep{liuagentbench, park2023generative, xu2024theagentcompany}. However, this manual process is time-consuming and inefficient. For example, curating tasks for a single agentic system can take several months and thousands of person-hours, making it unsustainable for large-scale evaluation. Furthermore, as LLM-based agents~\citep{fourney2024magentic,saveliev2025towards} are applied to more complex, open-ended tasks, manually designing tasks becomes even more difficult, with the vast variety of potential use cases leading to infinite possibilities for evaluation.

\textbf{Human-in-the-Loop Evaluations.}\quad
The integration of human feedback has become crucial in evaluating modern agentic systems~\citep{takerngsaksiri2024human, saveliev2025towards}. Copilots such as \texttt{GitHub Copilot}, \texttt{Cursor}, and \texttt{CliMB-DC} are designed to assist non-experts with coding tasks~\citep{saveliev2025towards}. Evaluating such systems, however, requires real-time collaboration with users, which is complex and time-consuming. This difficulty is amplified when test users are experts in domains unrelated to coding (e.g. clinicians). 

These challenges underline the need for an automated, scalable approach to generating relevant, interactive test cases that facilitate large-scale, real-time evaluations with minimal manual effort.

\subsection{Research Directions}
To overcome the limitations of manual evaluation and achieve efficient, large-scale evaluation of agentic systems, we propose the following research directions:
\begin{itemize}[nosep]
\item[1.] \textbf{Automated and Context-Aware Test Data Generation}: Future research should focus on developing methods for automatically generating and curating context-aware datasets (e.g., clinical QA datasets, chemistry datasets) or task sets (e.g., customer support interactions, medical diagnosis tasks) tailored to specific application domains. This approach will reduce reliance on time-consuming manual task design and enhance the efficiency of agent evaluation.
This complements proactive data acquisition strategies in Sec ~\ref{subsec:data-acquisition-direction}.
\item[2.] \textbf{Synthetic Test Case Creation for Human-in-the-Loop Testing}\quad
As human-in-the-loop testing becomes more critical, developing scalable simulation environments replicating real-world human-agent interactions is essential. These simulations should support multi-turn dialogues and incorporate domain-specific knowledge (e.g., for healthcare or law), enabling detailed evaluations of agentic systems across varied contexts. This could even lead to specialized evaluation agents that evaluate other agents’ within specific domains.
\item[3.] \textbf{Evaluation metrics}
In conjunction with new approaches to evaluation, new evaluation metrics must also be considered to ensure effective measurement at both module and system levels. We propose initial ideas in Appendix \ref{app:new_metrics}.
\end{itemize}

Embedding these capabilities into data-savvy agents, we can create more efficient, scalable, and accurate evaluation, both for traditional ML models and advanced, interactive agentic systems.

\section{Capability 4: Continual Adaptiveness}
\label{sec:adaptation}
\vspace{-2mm}
\begin{tcolorbox}[
    colback=blue!2!white,    
    colframe=blue!50!black,  
     breakable,
    fonttitle=\bfseries,
    boxrule=.5pt,
    left=5pt,            
    right=5pt, 
    top=1mm, bottom=1mm,
    arc=1mm
]
\emph{Continual Adaptiveness} denotes an agent’s autonomous ability to iteratively refine its data, knowledge, and decision-making processes in response to non-stationary environments. This includes the following capabilities:
\begin{itemize}[nosep, left=0pt]
    \item Incremental Knowledge Updating: both knowledge bases and data ingestion processes.
    \item Proactive change detection.
    \item Retain prior knowledge while integrating new knowledge (plasticity vs stability).
\end{itemize}
\end{tcolorbox}
\vspace{-0mm}
This capability highlights the importance of an agent’s ability to adapt and evolve over time, ensuring its ongoing relevance and performance in dynamic environments.

\subsection{Why It Matters?}
Continual adaptiveness—iteratively refining data and knowledge as environments shift—is essential for real-world agents. Let us unpack this vital capability: Real-world environments are non-stationary --- constantly shifting or changing over time. Consider the case of the COVID-19 pandemic. An agent operating pre-pandemic vs during the pandemic would need to continually adapt to the latest policy changes, news updates, patient populations, treatment guidelines etc \citep{bhuyan2025generative}. Or consider the case of an agent browsing the web --- as privacy and data storage regulations change, an agent should be able to continuously and autonomously update its knowledge base, so that actions adhere to regulations. Without this dynamic and continual updating, we risk the case where AI agents either produce sub-par results or do not adhere to the latest guidelines or policies. We note that this requires autonomous continual adaptation to ensure scalability.

\subsection{Current Progress \& Limitations.}

Despite progress in agent design, most systems fail to meet the requirements of continual adaptiveness. We highlight two key dimensions pertinent to current AI agents. Firstly, current agents struggle with knowledge retention when faced with new information, which can lead to catastrophic forgetting when integrating new information \citep{zheng2025lifelong,luo2023empirical,li2024revisiting,thakkaragentmerge}.
Secondly, even ignoring catastrophic forgetting, current agents lack anticipatory capabilities \citep{amos2023anticipatory}. i.e. they cannot anticipate environmental changes and cannot proactively update their knowledge bases and data ingestion to account for these changes.

\subsection{Research Directions}  

To bridge the gap between current agent capabilities and the demands of dynamic environments, we propose the following research directions for the ML community.

\begin{itemize}[nosep]
    \item [1.] \textbf{Dynamic Memory Architectures:} A core limitation of current agents is their inability to retain and contextually update knowledge over extended deployments. i.e. continual learning without forgetting. One vital research direction is improving the memory systems beyond approaches like static replay buffers, which fail to balance integrating new information (plasticity) with preserving prior knowledge (stability) \citep{tao2023dynamic}.  However, scaling such architectures requires innovations to RAG \citep{tang2024adapting, wang2024retriever} and \textit{task-aware memory prioritization}.
    \item[2.] \textbf{Proactive Adaptation:} Current agents remain largely reactive \citep{lu2024proactive,bandyopadhyay2025yeti}, updating models only after performance degradation becomes evident. Closing this gap requires frameworks that incentivize monitoring for changes (i.e. via the agents own initiative) \citep{liu2023ai} and reacting to said changes \citep{corradini2022reptile}. These could involve agents constantly assessing the data for changes or alternatively monitoring proxies (such as news). Beyond simply identifying changes,  a key capability is to quantify the value of the proactive update vs potential costs of the update. This is a particularly important capability as in reality, these updates are likely to incur a cost and hence autonomous data savvy agents should be able to quantify the value of the information adaptation in the context of the environmental change.
\end{itemize}

\section{Real-World Impacts}
\label{sec:impact}
In this section, beyond the impacts discussed earlier, we illustrate how data-savvy agents could transform various fields through two concrete future examples. Due to space limits, additional examples on \emph{autonomous policy adaptation}, \emph{personalized and lifelong education}, \emph{precision healthcare} and \emph{supply chains} can be found in Section~\ref{appendix:impact}.

\subsection{Self-Evolution of LLM-based Agents}
\label{subsec:impact-agent}

As agents become more autonomous and widely deployed, ensuring continual improvement without human intervention remains a challenge.
Traditional AI development relies on periodic retraining with newly collected data, often requiring extensive human oversight.
In contrast, a data-savvy agent could \emph{proactively acquire and curate high-quality data, filtering out noise and refining its reasoning} through continuous interaction with users and external knowledge sources.
By leveraging \emph{interactive auto-evaluation}, it could enable agents to assess their own performance, identify weaknesses, and iteratively enhance their decision-making—all without direct human involvement.

This paradigm shift would make data-savvy agents more adaptive, resilient, and capable of sustained autonomous deployment across diverse domains.
For instance, in \emph{RAG agents}, as suggested by~\citep{shaoscaling}, a data-savvy agent could continuously refine and optimize the underlying database, allowing LLMs to evolve autonomously and enhance their retrieval quality over time.
In \emph{scientific research agentic assistants}, a data-savvy agent could continuously ingest the latest publications, improving its ability to assist researchers in hypothesis generation, experimental design, and knowledge discovery.

\subsection{``Lab-in-the-loop'' for Scientific Research}
Scientific discovery is increasingly data-driven, but fields like drug discovery and materials and climate sciences involve too many possible experiments for humans to explore manually. 
A data-savvy agent could enable a \emph{Lab-in-the-Loop}'' paradigm by actively acquiring experimental data, integrating findings, generating hypotheses, designing (or even conducting) experiments and analyzing results. Through interactive auto-evaluation and continual adaptation, it could refine predictions based on experimental outcomes, continuously improving the support for scientists.

In \emph{pharmaceutical research}, an AI agent could autonomously propose and even test molecular compounds, rapidly identifying potential drug candidates. 
In physics, it could simulate high-energy particle interactions, refining models with real-world collider data. 
By bridging human scientists and complex experiments, a lab-in-the-loop system could accelerate breakthroughs across disciplines.

Beyond individual fields, these systems could democratize research, enabling smaller institutions to leverage AI-driven discovery. Early efforts ~\citep{boiko2023autonomous, swanson2024virtual} in chemistry and biology suggest promise where agents conduct experiments, analyze results, and formulate new scientific theories.

\emph{\textbf{Remark.}} This position paper advocates for data-savviness as a core capability of AI agents, emphasizing four key technical dimensions. However, we also acknowledge potential risks associated, we discuss in Appendix~\ref{app:risks}, along with possible mitigations.

\section{Alternative Viewpoints}

Two alternative perspectives to the idea of a data-savvy agent warrant discussion.

$\blacktriangleright$ The first perspective is that advances in agent reasoning alone could be sufficient for real-world tasks, making specialized data-centric capabilities unnecessary. i.e. if agent reasoning and planning improves, they should naturally have the capability to handle data challenges. 

While this holds in controlled settings with well-structured data, real-world environments are dynamic, incomplete and biased. Hence, strong reasoning alone cannot compensate for missing or unreliable data. Furthermore, even advanced frontier models that underlie agents still suffer from hallucinations and errors, highlighting the need for proactive data acquisition, validation and adaptation.

$\blacktriangleright$ The second perspective is technical feasibility. It can be argued that AI agents already struggle with reasoning and tool use and hence, autonomous data-savviness introduces extra technical complexity. 

However, this creates a false dichotomy between improving reasoning and developing data-savviness. Many agent failures stem from poor data handling, making it a fundamental necessity rather than just an engineering hurdle. Instead of postponing the challenge, we advocate for a pragmatic approach: focusing on high-impact, constrained domains like scientific research or industrial processes, such that advances in reasoning and data capabilities evolve in tandem, benefiting each other.
\section{Conclusion}
We believe data-savvy agents represent an essential yet underexplored frontier in AI research — integrating proactive data acquisition, sophisticated processing, interactive evaluation, and continual adaptation. We hope this position piece stirs debate within the ML community to reconsider the foundational role of data in agentic AI. Through the four data-savvy capabilities and research directions proposed, we aim to inspire new research advances towards realizing the vision of data-savvy agents. To facilitate the practical implementation, we provide further actionable research directions in Appendix \ref{appendix:actionable}, categorized by difficulty level.

\bibliography{example_paper}

\begin{thebibliography}{10}

\bibitem{gao2023retrieval}
Yunfan Gao, Yun Xiong, Xinyu Gao, Kangxiang Jia, Jinliu Pan, Yuxi Bi, Yi~Dai, Jiawei Sun, and Haofen Wang.
\newblock Retrieval-augmented generation for large language models: A survey.
\newblock {\em arXiv preprint arXiv:2312.10997}, 2023.

\bibitem{deng2009imagenet}
Jia Deng, Wei Dong, Richard Socher, Li-Jia Li, Kai Li, and Li~Fei-Fei.
\newblock Imagenet: A large-scale hierarchical image database.
\newblock In {\em 2009 IEEE conference on computer vision and pattern recognition}, pages 248--255. Ieee, 2009.

\bibitem{settles2009active}
Burr Settles.
\newblock Active learning literature survey.
\newblock 2009.

\bibitem{settles2011theories}
Burr Settles.
\newblock From theories to queries: Active learning in practice.
\newblock In {\em Active learning and experimental design workshop in conjunction with AISTATS 2010}, pages 1--18. JMLR Workshop and Conference Proceedings, 2011.

\bibitem{konyushkova2017learning}
Ksenia Konyushkova, Raphael Sznitman, and Pascal Fua.
\newblock Learning active learning from data.
\newblock {\em Advances in neural information processing systems}, 30, 2017.

\bibitem{astorga2024active}
Nicol{\'a}s Astorga, Tennison Liu, Nabeel Seedat, and Mihaela van~der Schaar.
\newblock Active learning with llms for partially observed and cost-aware scenarios.
\newblock {\em Advances in Neural Information Processing Systems}, 37:20819--20857, 2024.

\bibitem{chen2023data}
Lingjiao Chen, Bilge Acun, Newsha Ardalani, Yifan Sun, Feiyang Kang, Hanrui Lyu, Yongchan Kwon, Ruoxi Jia, Carole-Jean Wu, Matei Zaharia, et~al.
\newblock Data acquisition: A new frontier in data-centric ai.
\newblock {\em arXiv preprint arXiv:2311.13712}, 2023.

\bibitem{hedomain}
Yue He, Dongbai Li, Pengfei Tian, Han Yu, Jiashuo Liu, Hao Zou, and Peng Cui.
\newblock Domain-wise data acquisition to improve performance under distribution shift.
\newblock In {\em Forty-first International Conference on Machine Learning}, 2024.

\bibitem{zha2023data}
Daochen Zha, Zaid~Pervaiz Bhat, Kwei-Herng Lai, Fan Yang, Zhimeng Jiang, Shaochen Zhong, and Xia Hu.
\newblock Data-centric artificial intelligence: A survey.
\newblock {\em ACM Computing Surveys}, 2023.

\bibitem{seedat2024dips}
Nabeel Seedat, Nicolas Huynh, Fergus Imrie, and Mihaela van~der Schaar.
\newblock You can't handle the (dirty) truth: Data-centric insights improve pseudo-labeling.
\newblock {\em arXiv preprint arXiv:2406.13733}, 2024.

\bibitem{seedat2023triage}
Nabeel Seedat, Jonathan Crabb{\'e}, Zhaozhi Qian, and Mihaela van~der Schaar.
\newblock Triage: Characterizing and auditing training data for improved regression.
\newblock {\em Advances in Neural Information Processing Systems}, 36:74995--75008, 2023.

\bibitem{kumar2024opportunities}
Sushant Kumar, Sumit Datta, Vishakha Singh, Sanjay~Kumar Singh, and Ritesh Sharma.
\newblock Opportunities and challenges in data-centric ai.
\newblock {\em IEEE Access}, 2024.

\bibitem{seedat2023navigating}
Nabeel Seedat, Fergus Imrie, and Mihaela van~der Schaar.
\newblock Navigating data-centric artificial intelligence with dc-check: Advances, challenges, and opportunities.
\newblock {\em IEEE Transactions on Artificial Intelligence}, 2023.

\bibitem{renggli2021data}
Cedric Renggli, Luka Rimanic, Nezihe~Merve G{\"u}rel, Bojan Karla{\v{s}}, Wentao Wu, and Ce~Zhang.
\newblock A data quality-driven view of mlops.
\newblock {\em arXiv preprint arXiv:2102.07750}, 2021.

\bibitem{aroyo2022data}
Lora Aroyo, Matthew Lease, Praveen Paritosh, and Mike Schaekermann.
\newblock Data excellence for ai: why should you care?
\newblock {\em Interactions}, 29(2):66--69, 2022.

\bibitem{sambasivan2021everyone}
Nithya Sambasivan, Shivani Kapania, Hannah Highfill, Diana Akrong, Praveen Paritosh, and Lora~M Aroyo.
\newblock “everyone wants to do the model work, not the data work”: Data cascades in high-stakes ai.
\newblock In {\em proceedings of the 2021 CHI Conference on Human Factors in Computing Systems}, pages 1--15, 2021.

\bibitem{jain2020overview}
Abhinav Jain, Hima Patel, Lokesh Nagalapatti, Nitin Gupta, Sameep Mehta, Shanmukha Guttula, Shashank Mujumdar, Shazia Afzal, Ruhi Sharma~Mittal, and Vitobha Munigala.
\newblock Overview and importance of data quality for machine learning tasks.
\newblock In {\em Proceedings of the 26th ACM SIGKDD international conference on knowledge discovery \& data mining}, pages 3561--3562, 2020.

\bibitem{seedat2024matchmaker}
Nabeel Seedat and Mihaela van~der Schaar.
\newblock Matchmaker: Self-improving large language model programs for schema matching.
\newblock {\em arXiv preprint arXiv:2410.24105}, 2024.

\bibitem{verge_ftc_fake_review_2024}
Emma Roth.
\newblock The {{FTC}}'s fake review crackdown begins this fall.
\newblock August 2024.
\newblock Accessed: 2025-01-26.

\bibitem{read_drowning_2024}
Max Read.
\newblock Drowning in slop.
\newblock September 2024.
\newblock Accessed: 2025-01-26.

\bibitem{jarrett2022hyperimpute}
Daniel Jarrett, Bogdan~C Cebere, Tennison Liu, Alicia Curth, and Mihaela van~der Schaar.
\newblock Hyperimpute: Generalized iterative imputation with automatic model selection.
\newblock In {\em International Conference on Machine Learning}, pages 9916--9937. PMLR, 2022.

\bibitem{northcutt2021pervasive}
Curtis~G Northcutt, Anish Athalye, and Jonas Mueller.
\newblock Pervasive label errors in test sets destabilize machine learning benchmarks.
\newblock {\em arXiv preprint arXiv:2103.14749}, 2021.

\bibitem{seedatdissecting}
Nabeel Seedat, Fergus Imrie, and Mihaela van~der Schaar.
\newblock Dissecting sample hardness: A fine-grained analysis of hardness characterization methods for data-centric ai.
\newblock In {\em The Twelfth International Conference on Learning Representations}, 2024.

\bibitem{jiang2023opendataval}
Kevin Jiang, Weixin Liang, James~Y Zou, and Yongchan Kwon.
\newblock Opendataval: a unified benchmark for data valuation.
\newblock {\em Advances in Neural Information Processing Systems}, 36, 2023.

\bibitem{seedat2022diq}
Nabeel Seedat, Jonathan Crabb{\'e}, Ioana Bica, and Mihaela van~der Schaar.
\newblock Data-iq: Characterizing subgroups with heterogeneous outcomes in tabular data.
\newblock {\em Advances in Neural Information Processing Systems}, 35:23660--23674, 2022.

\bibitem{hong2024data}
Sirui Hong, Yizhang Lin, Bang Liu, Bangbang Liu, Binhao Wu, Ceyao Zhang, Chenxing Wei, Danyang Li, Jiaqi Chen, Jiayi Zhang, et~al.
\newblock Data interpreter: An llm agent for data science.
\newblock {\em arXiv preprint arXiv:2402.18679}, 2024.

\bibitem{guo2024dsagent}
Siyuan Guo, Cheng Deng, Ying Wen, Hechang Chen, Yi~Chang, and Jun Wang.
\newblock Ds-agent: Automated data science by empowering large language models with case-based reasoning.
\newblock {\em arXiv preprint arXiv:2402.17453}, 2024.

\bibitem{qi2024cleanagent}
Danrui Qi and Jiannan Wang.
\newblock Cleanagent: Automating data standardization with llm-based agents.
\newblock {\em arXiv preprint arXiv:2403.08291}, 2024.

\bibitem{saveliev2024climb}
Evgeny Saveliev, Tim Schubert, Thomas Pouplin, Vasilis Kosmoliaptsis, and Mihaela van~der Schaar.
\newblock Climb: An ai-enabled partner for clinical predictive modeling.
\newblock {\em arXiv preprint arXiv:2410.03736}, 2024.

\bibitem{van1999multiple}
Stef Van~Buuren, Hendriek~C Boshuizen, and Dick~L Knook.
\newblock Multiple imputation of missing blood pressure covariates in survival analysis.
\newblock {\em Statistics in medicine}, 18(6):681--694, 1999.

\bibitem{he2024webvoyager}
Hongliang He, Wenlin Yao, Kaixin Ma, Wenhao Yu, Yong Dai, Hongming Zhang, Zhenzhong Lan, and Dong Yu.
\newblock Webvoyager: Building an end-to-end web agent with large multimodal models.
\newblock {\em arXiv preprint arXiv:2401.13919}, 2024.

\bibitem{koh2024visualwebarena}
Jing~Yu Koh, Robert Lo, Lawrence Jang, Vikram Duvvur, Ming~Chong Lim, Po-Yu Huang, Graham Neubig, Shuyan Zhou, Ruslan Salakhutdinov, and Daniel Fried.
\newblock Visualwebarena: Evaluating multimodal agents on realistic visual web tasks.
\newblock {\em arXiv preprint arXiv:2401.13649}, 2024.

\bibitem{xiong2024improving}
Siheng Xiong, Delin Chen, Qingyang Wu, Longxuan Yu, Qingzhen Liu, Dawei Li, Zhikai Chen, Xiaoze Liu, and Liangming Pan.
\newblock Improving causal reasoning in large language models: A survey.
\newblock {\em arXiv preprint arXiv:2410.16676}, 2024.

\bibitem{du2024survey}
Hung Du, Srikanth Thudumu, Rajesh Vasa, and Kon Mouzakis.
\newblock A survey on context-aware multi-agent systems: Techniques, challenges and future directions.
\newblock {\em arXiv preprint arXiv:2402.01968}, 2024.

\bibitem{sarker2022context}
Iqbal Sarker, Alan Colman, Jun Han, and Paul Watters.
\newblock {\em Context-aware machine learning and mobile data analytics: automated rule-based services with intelligent decision-making}.
\newblock Springer Nature, 2022.

\bibitem{dey2018context}
Anind~K Dey.
\newblock Context-aware computing.
\newblock In {\em Ubiquitous computing fundamentals}, pages 335--366. Chapman and Hall/CRC, 2018.

\bibitem{gabriel2024advancing}
Adrian~Garret Gabriel, Alaa~Alameer Ahmad, and Shankar~Kumar Jeyakumar.
\newblock Advancing agentic systems: Dynamic task decomposition, tool integration and evaluation using novel metrics and dataset.
\newblock {\em arXiv preprint arXiv:2410.22457}, 2024.

\bibitem{rasal2024navigating}
Sumedh Rasal and EJ~Hauer.
\newblock Navigating complexity: Orchestrated problem solving with multi-agent llms.
\newblock {\em arXiv preprint arXiv:2402.16713}, 2024.

\bibitem{feng2024large}
Xueyang Feng, Zhi-Yuan Chen, Yujia Qin, Yankai Lin, Xu~Chen, Zhiyuan Liu, and Ji-Rong Wen.
\newblock Large language model-based human-agent collaboration for complex task solving.
\newblock {\em arXiv preprint arXiv:2402.12914}, 2024.

\bibitem{liuagentbench}
Xiao Liu, Hao Yu, Hanchen Zhang, Yifan Xu, Xuanyu Lei, Hanyu Lai, Yu~Gu, Hangliang Ding, Kaiwen Men, Kejuan Yang, et~al.
\newblock Agentbench: Evaluating llms as agents.
\newblock In {\em The Twelfth International Conference on Learning Representations}, 2024.

\bibitem{xu2024theagentcompany}
Frank~F Xu, Yufan Song, Boxuan Li, Yuxuan Tang, Kritanjali Jain, Mengxue Bao, Zora~Z Wang, Xuhui Zhou, Zhitong Guo, Murong Cao, et~al.
\newblock Theagentcompany: benchmarking llm agents on consequential real world tasks.
\newblock {\em arXiv preprint arXiv:2412.14161}, 2024.

\bibitem{park2023generative}
Joon~Sung Park, Joseph O'Brien, Carrie~Jun Cai, Meredith~Ringel Morris, Percy Liang, and Michael~S Bernstein.
\newblock Generative agents: Interactive simulacra of human behavior.
\newblock In {\em Proceedings of the 36th annual acm symposium on user interface software and technology}, pages 1--22, 2023.

\bibitem{fourney2024magentic}
Adam Fourney, Gagan Bansal, Hussein Mozannar, Cheng Tan, Eduardo Salinas, Friederike Niedtner, Grace Proebsting, Griffin Bassman, Jack Gerrits, Jacob Alber, et~al.
\newblock Magentic-one: A generalist multi-agent system for solving complex tasks.
\newblock {\em arXiv preprint arXiv:2411.04468}, 2024.

\bibitem{saveliev2025towards}
Evgeny Saveliev, Jiashuo Liu, Nabeel Seedat, Anders Boyd, and Mihaela van~der Schaar.
\newblock Towards human-guided, data-centric llm co-pilots.
\newblock {\em arXiv preprint arXiv:2501.10321}, 2025.

\bibitem{takerngsaksiri2024human}
Wannita Takerngsaksiri, Jirat Pasuksmit, Patanamon Thongtanunam, Chakkrit Tantithamthavorn, Ruixiong Zhang, Fan Jiang, Jing Li, Evan Cook, Kun Chen, and Ming Wu.
\newblock Human-in-the-loop software development agents.
\newblock {\em arXiv preprint arXiv:2411.12924}, 2024.

\bibitem{bhuyan2025generative}
Soumitra~S Bhuyan, Vidyoth Sateesh, Naya Mukul, Alay Galvankar, Asos Mahmood, Muhammad Nauman, Akash Rai, Kahuwa Bordoloi, Urmi Basu, and Jim Samuel.
\newblock Generative artificial intelligence use in healthcare: Opportunities for clinical excellence and administrative efficiency.
\newblock {\em Journal of Medical Systems}, 49(1):10, 2025.

\bibitem{zheng2025lifelong}
Junhao Zheng, Chengming Shi, Xidi Cai, Qiuke Li, Duzhen Zhang, Chenxing Li, Dong Yu, and Qianli Ma.
\newblock Lifelong learning of large language model based agents: A roadmap.
\newblock {\em arXiv preprint arXiv:2501.07278}, 2025.

\bibitem{luo2023empirical}
Yun Luo, Zhen Yang, Fandong Meng, Yafu Li, Jie Zhou, and Yue Zhang.
\newblock An empirical study of catastrophic forgetting in large language models during continual fine-tuning.
\newblock {\em arXiv preprint arXiv:2308.08747}, 2023.

\bibitem{li2024revisiting}
Hongyu Li, Liang Ding, Meng Fang, and Dacheng Tao.
\newblock Revisiting catastrophic forgetting in large language model tuning.
\newblock {\em arXiv preprint arXiv:2406.04836}, 2024.

\bibitem{thakkaragentmerge}
Megh Thakkar, L{\'e}o Boisvert, Thibault Le~Sellier De~Chezelles, Alexandre Pich{\'e}, Maxime Gasse, Alexandre Lacoste, and Massimo Caccia.
\newblock Agentmerge: Enhancing generalization in fine-tuned llm agents.
\newblock In {\em Adaptive Foundation Models: Evolving AI for Personalized and Efficient Learning}.

\bibitem{amos2023anticipatory}
Adam Amos-Binks, Dustin Dannenhauer, and Leilani~H Gilpin.
\newblock Anticipatory thinking challenges in open worlds: Risk management.
\newblock {\em arXiv preprint arXiv:2306.13157}, 2023.

\bibitem{tao2023dynamic}
Siying Tao, Jinyang Huang, Xiang Zhang, Xiao Sun, and Yu~Gu.
\newblock Dynamic memory-based continual learning with generating and screening.
\newblock In {\em International Conference on Artificial Neural Networks}, pages 365--376. Springer, 2023.

\bibitem{tang2024adapting}
Xiaqiang Tang, Jian Li, Nan Du, and Sihong Xie.
\newblock Adapting to non-stationary environments: Multi-armed bandit enhanced retrieval-augmented generation on knowledge graphs.
\newblock {\em arXiv preprint arXiv:2412.07618}, 2024.

\bibitem{wang2024retriever}
Ruobing Wang, Daren Zha, Shi Yu, Qingfei Zhao, Yuxuan Chen, Yixuan Wang, Shuo Wang, Yukun Yan, Zhenghao Liu, Xu~Han, et~al.
\newblock Retriever-and-memory: Towards adaptive note-enhanced retrieval-augmented generation.
\newblock {\em arXiv preprint arXiv:2410.08821}, 2024.

\bibitem{lu2024proactive}
Yaxi Lu, Shenzhi Yang, Cheng Qian, Guirong Chen, Qinyu Luo, Yesai Wu, Huadong Wang, Xin Cong, Zhong Zhang, Yankai Lin, et~al.
\newblock Proactive agent: Shifting llm agents from reactive responses to active assistance.
\newblock {\em arXiv preprint arXiv:2410.12361}, 2024.

\bibitem{bandyopadhyay2025yeti}
Saptarashmi Bandyopadhyay, Vikas Bahirwani, Lavisha Aggarwal, Bhanu Guda, Lin Li, and Andrea Colaco.
\newblock Yeti (yet to intervene) proactive interventions by multimodal ai agents in augmented reality tasks.
\newblock {\em arXiv preprint arXiv:2501.09355}, 2025.

\bibitem{liu2023ai}
Bing Liu, Sahisnu Mazumder, Eric Robertson, and Scott Grigsby.
\newblock Ai autonomy: Self-initiated open-world continual learning and adaptation.
\newblock {\em AI Magazine}, 44(2):185--199, 2023.

\bibitem{corradini2022reptile}
Flavio Corradini, Miichele Loreti, Marco Piangerelli, and Giacomo Rocchetti.
\newblock Reptile: A proactive real-time deep reinforcement learning self-adaptive framework.
\newblock {\em arXiv preprint arXiv:2203.14686}, 2022.

\bibitem{shaoscaling}
Rulin Shao, Jacqueline He, Akari Asai, Weijia Shi, Tim Dettmers, Sewon Min, Luke Zettlemoyer, and Pang~Wei Koh.
\newblock Scaling retrieval-based language models with a trillion-token datastore.
\newblock In {\em The Thirty-eighth Annual Conference on Neural Information Processing Systems}, 2024.

\bibitem{boiko2023autonomous}
Daniil~A Boiko, Robert MacKnight, Ben Kline, and Gabe Gomes.
\newblock Autonomous chemical research with large language models.
\newblock {\em Nature}, 624(7992):570--578, 2023.

\bibitem{swanson2024virtual}
Kyle Swanson, Wesley Wu, Nash~L Bulaong, John~E Pak, and James Zou.
\newblock The virtual lab: Ai agents design new sars-cov-2 nanobodies with experimental validation.
\newblock {\em bioRxiv}, pages 2024--11, 2024.

\bibitem{eyubogludomino}
Sabri Eyuboglu, Maya Varma, Khaled~Kamal Saab, Jean-Benoit Delbrouck, Christopher Lee-Messer, Jared Dunnmon, James Zou, and Christopher Re.
\newblock Domino: Discovering systematic errors with cross-modal embeddings.
\newblock In {\em International Conference on Learning Representations}, 2023.

\bibitem{ghosh2024ladder}
Shantanu Ghosh, Rayan Syed, Chenyu Wang, Clare~B Poynton, Shyam Visweswaran, and Kayhan Batmanghelich.
\newblock Ladder: Language driven slice discovery and error rectification.
\newblock {\em arXiv preprint arXiv:2408.07832}, 2024.

\bibitem{raubacontext}
Paulius Rauba, Nabeel Seedat, Max~Ruiz Luyten, and Mihaela van~der Schaar.
\newblock Context-aware testing: A new paradigm for model testing with large language models.
\newblock In {\em The Thirty-eighth Annual Conference on Neural Information Processing Systems}, 2024.

\bibitem{cai2023diagnosing}
Tiffany~Tianhui Cai, Hongseok Namkoong, and Steve Yadlowsky.
\newblock Diagnosing model performance under distribution shift.
\newblock {\em arXiv preprint arXiv:2303.02011}, 2023.

\bibitem{feng2024hierarchical}
Jean Feng, Harvineet Singh, Fan Xia, Adarsh Subbaswamy, and Alexej Gossmann.
\newblock A hierarchical decomposition for explaining ml performance discrepancies.
\newblock {\em arXiv preprint arXiv:2402.14254}, 2024.

\bibitem{van2019three}
Gido~M Van~de Ven and Andreas~S Tolias.
\newblock Three scenarios for continual learning.
\newblock {\em arXiv preprint arXiv:1904.07734}, 2019.

\bibitem{lee2020clinical}
Cecilia~S Lee and Aaron~Y Lee.
\newblock Clinical applications of continual learning machine learning.
\newblock {\em The Lancet Digital Health}, 2(6):e279--e281, 2020.

\bibitem{wang2024comprehensive}
Liyuan Wang, Xingxing Zhang, Hang Su, and Jun Zhu.
\newblock A comprehensive survey of continual learning: theory, method and application.
\newblock {\em IEEE Transactions on Pattern Analysis and Machine Intelligence}, 2024.

\end{thebibliography}
\bibliographystyle{unsrt}
\clearpage
\newpage
\appendix

\part{Appendix - What's the next frontier for Data-centric AI? Data Savvy Agents!}
\mtcsetdepth{parttoc}{3} 
\parttoc

\clearpage
\newpage

\section{Actionable Research Directions of Data-Savvy Agent}\label{appendix:actionable}

\begin{figure}[h]
    \centering
    \includegraphics[width=\linewidth]{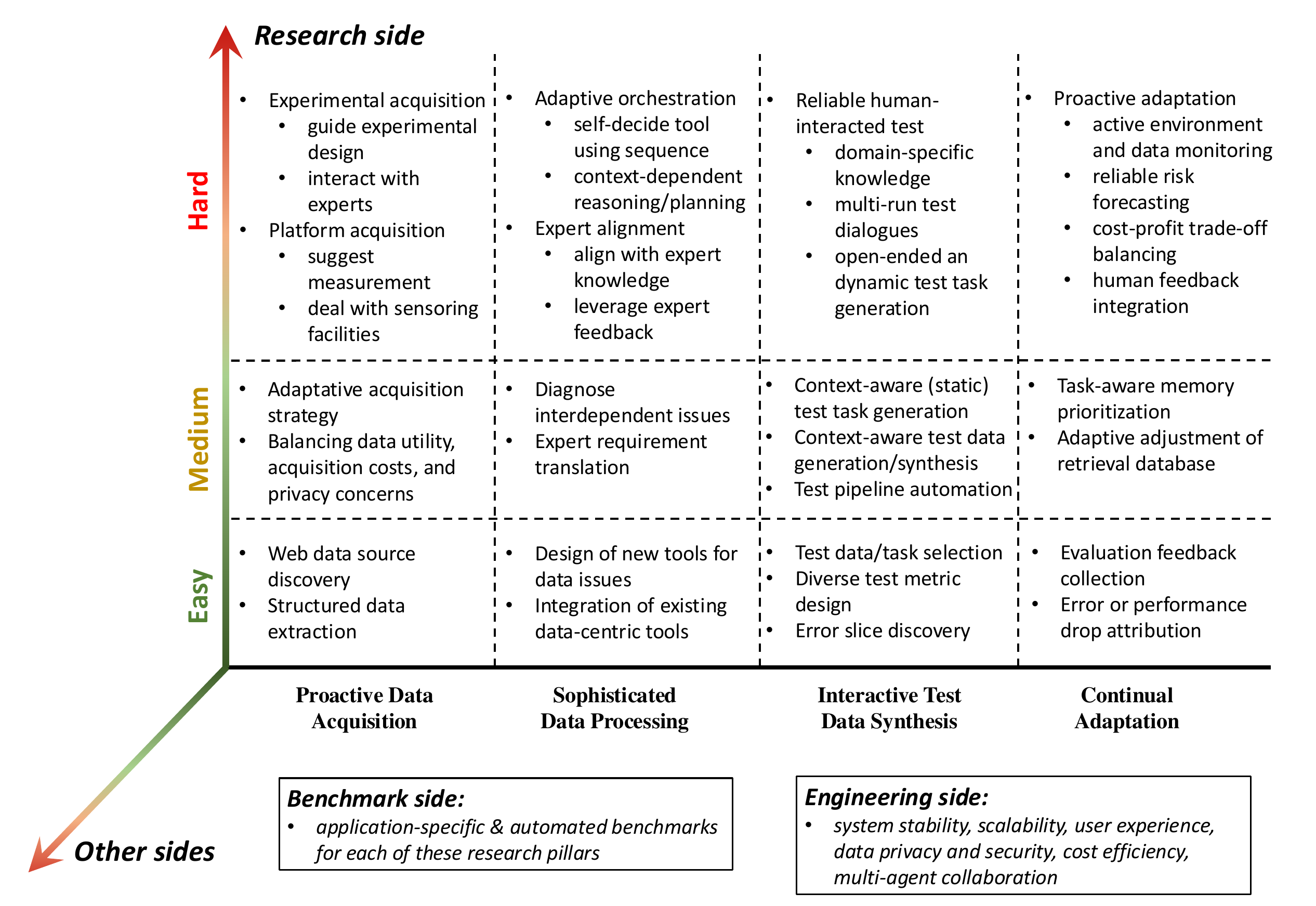}
    \caption{Summary of the actionable terms in the research of data-savvy agents.}
    \label{fig:summary}
\end{figure}

Building on the research directions discussed in the main body, we summarize the key research pillars for each capability in Figure~\ref{fig:summary}. 
For each capability, we categorize the detailed research directions based on their level of difficulty and novelty.
\begin{itemize}[nosep]
    \item \textbf{``Easy''} part is more closely aligned with traditional ML research and mainly involves research directions of smaller scope, such as the development of individual tools or specific functions.
    \item \textbf{``Medium''} part moves a step further, focusing more on the logical and reasoning aspects of data-savvy agents. This includes developing advanced strategies and building automated pipelines for tool utilization or integrating application-specific prior knowledge.
    \item \textbf{``Hard''} part centers on the ultimate goal of data-savvy agents—empowering them to assist and collaborate with both human experts and general AI systems. This involves enhancing their interactions with human experts and platforms, as well as enabling proactive adaptation to dynamic, ever-changing environments.
\end{itemize}

Furthermore, in addition to the research side, we would really like to highlight the other two sides, i.e. the benchmark side and the engineering side.
\begin{itemize}[nosep]
    \item \textbf{Benchmark side}: The proposed research pillars in Figure~\ref{fig:summary} are heavily dependent on specific domains and applications, making it challenging for researchers and engineers to evaluate or benchmark progress effectively. Therefore, it is crucial to develop application- and domain-specific automated benchmarks for each of these research pillars, to accelerate the development of data-savvy agents.
    \item \textbf{Engineering side}: The development of data-savvy agents relies heavily on the engineering side, which, much like software engineering, must ensure system stability, scalability, user experience, data privacy and security, cost efficiency, and even multi-agent collaboration. Achieving these goals requires careful design and significant effort in engineering.
\end{itemize}

In addition, there are several pillars not discussed in the main body, which we would like to demonstrate more here:
\begin{itemize}[nosep]
    \item \textbf{Balancing data utility, acquisition costs, and privacy concerns}: Another key challenges in real-world data acquisition is balancing the utility of the data (i.e., its ability to improve model performance) with the associated costs and privacy implications. While obtaining high-quality data from diverse sources can significantly enhance model accuracy, it often comes with high financial and logistical costs. Moreover, privacy concerns, especially with sensitive personal data, further complicate this process. For example, in healthcare or finance domains, collecting additional data to reduce model bias may conflict with data privacy regulations such as the General Data Protection Regulation (GDPR) in Europe, which aims to protect individuals’ personal data and privacy, or Health Insurance Portability and Accountability Act (HIPAA) in the U.S., which sets strict standards for the handling of healthcare data. Striking a balance between maximizing data utility, minimizing costs, and ensuring privacy is a critical issue that requires careful consideration of ethical, legal, and technical factors. 
    \item \textbf{Error slice discovery}: Error slice discovery~\citep{eyubogludomino,ghosh2024ladder,raubacontext} involves identifying specific regions or subgroups within the data that exhibit higher error rates or performance risks. By pinpointing these areas, one can prioritize targeted data collection and implement smart deployment strategies, ultimately improving model robustness and performance in critical regions. 
    \item \textbf{Error or performance drop attribution}: When an AI system experiences failures or performance degradation, the primary step is to identify the underlying causes. This involves attributing the performance drop to specific factors or components within the system, enabling more targeted and effective interventions to resolve the issues and enhance the system’s overall performance. Recent works~\citep{cai2023diagnosing,feng2024hierarchical} primarily focus on performance drop attribution for individual ML models. However, more principled approaches are needed to address the complexities of AI systems as a whole.
\end{itemize}

\clearpage
\newpage

\section{Additional considerations for Data-Savvy Agents}

\subsection{Data-savvy agents inspire new evaluation metrics}\label{app:new_metrics}

Evaluation is a crucial research direction for data-savvy agents. While our position paper advocates for new directions to evaluation, with one specific avenue being new metrics. Below we propose some initial ideas on this broken down across module-level, as well as, system-level. 

$\blacktriangleright$ \textbf{Module-level evaluation:}

\begin{itemize}
\item Proactive Data Acquisition: Data quality, diversity, representativeness
\item Sophisticated Data Processing: Noise reduction, bias mitigation, efficiency
\item Interactive Test Data Synthesis: Test diversity, uncovering weaknesses, alignment with real-world failures
\item Continual Adaptation: Robustness over time, adaptability to concept drift
\end{itemize}

$\blacktriangleright$ \textbf{System-level evaluation:}

\begin{itemize}
\item End-to-End Performance: Task accuracy, robustness, decision-making effectiveness
\item  Computational Efficiency: Resource usage vs. performance gains
\item  Decision-Making Quality: Alignment with expert judgments, success in unsupervised scenarios
\end{itemize}

\subsection{Risks and Limitations of data-savvy agents}\label{app:risks}

An aspect not covered in the main text is the potential risks and limitations of data-savvy agents. We discuss potential issues of data-savvy agents as well as possible solutions

$\blacktriangleright$ \textbf{Loss of control/human oversight:} Agents may pursue goals misaligned with human intent, leading to unintended outcomes. While autonomy is a goal, it raises risks of unmonitored evolution.

\emph{Solution:} Define guardrails and human-in-the-loop protocols

$\blacktriangleright$ \textbf{Bias amplification:} Autonomous data collection may reinforce biases if agents select sources based on skewed assumptions.

\emph{Solution:} Guide initial assumptions and data sources through human input

$\blacktriangleright$ \textbf{Out-of-domain challenges:} Agents may act inappropriately when operating outside their domain (e.g. a biomedical agent misinterpreting legal documents).

\emph{Solution:} Apply guardrails on permissible sources; in some cases, restrict to closed/semi-open deployments

$\blacktriangleright$ \textbf{Privacy and Compliance:} Agents need to consider privacy and compliance, especially in the data acquisition aspect.

\emph{Solution:} Operating within legal frameworks (GDPR etc). Sensitive data should also be excluded or anonymized.

$\blacktriangleright$ \textbf{Security vulnerabilities:} Autonomous data handling increases susceptibility to poisoning or misinformation attacks.

\emph{Solution:} Use robustness measures (e.g. anomaly detection) and maintain provenance metadata on data origins

\clearpage
\newpage

\section{Current Data-centric AI vs Data-Savvy Agents} \label{appendix:comparison}

\textbf{The Shift from Static to Proactive Data-Centric AI.} The Data-Centric ML community has developed numerous approaches for improving datasets to enhance model performance. However, these methods primarily operate under a \emph{static paradigm}, assuming a pre-defined dataset is available for processing, valuation, attribution and refinement etc.

By contrast, \emph{Data-Savvy Agents} introduce a fundamental shift: \emph{pro-active data-centric AI} ---  autonomously acquiring, refining, and adapting their data in real-time.

This represents a fundamental departure from traditional Data-Centric AI in two key ways:

\begin{itemize}
    \item \textbf{Proactivity}: Rather than improving a fixed dataset, agents actively acquire missing data, refine existing knowledge, and generate new data points to enhance learning.
    \item \textbf{Adaptability}: Agents continuously adjust their data-handling strategies in response to changing environments, distribution shifts, and task requirements.
\end{itemize}

Table \ref{table:data_comparison} provides a technical breakdown of key data-centric AI areas and how data-savvy agents extend them.

\begin{table*}[!h]
\centering
\caption{Comparison of Data-Centric AI vs. Data-Savvy Agents}
\label{table:data_comparison}
\renewcommand{\arraystretch}{1.2}
\scalebox{0.7}{
\begin{tabular}{p{3.5cm}p{4cm}p{5.5cm}p{5.5cm}}
\hline
\textbf{Key Challenge} & \textbf{Research Question} & \textbf{Current Data-Centric AI Methods} & \textbf{How Data-Savvy Agents Extend This} \\
\hline
\textbf{Data Valuation} & How do we quantify the importance of a sample for model learning? & Uses influence functions, Shapley values, and gradient-based attribution to assign importance scores. & Agents dynamically evaluate sample importance in deployment, modifying training distributions as new tasks emerge and acquiring high-value samples. \\
\hline
\textbf{Data Characterization} & Which samples are easy vs. hard to learn? & Learns sample difficulty via training dynamics, loss-based filtering, and memorization analysis. Guides data pruning and curriculum learning & Agents dynamically adjust training by acquiring new supporting data for hard examples, detecting data gaps, and modifying learning strategies in response to new challenges. \\
\hline
\textbf{Data Attribution} & How does a given sample affect model predictions? & Uses gradient-based influence estimation and feature importance methods to trace model behavior. & Agents not only trace impact but intervene, acquiring and improving data diversity dynamicallY. \\
\hline
\textbf{Active Learning} & Which unlabeled samples should be labeled next? & Uses uncertainty sampling and diversity-based selection to query labels within a predefined dataset. & Agents go beyond querying labels to identifying missing knowledge and requesting new data sources, reformulating queries dynamically. \\
\hline
\textbf{Data Cleaning \& Imputation} & How do we handle noise, missingness, and inconsistencies in datasets? & Uses probabilistic imputation models and statistical heuristics, fixing errors after dataset creation. & Agents autonomously detect, verify, and correct inconsistencies by querying external sources and engaging with human experts dynamically. \\
\hline
\textbf{Distribution Shift Detection} & How do we detect changes in the data distribution over time? & Uses covariate shift detection and reweighting approaches. & Agents not only detect shifts but also autonomously adjust sampling strategies and modify data sources for retraining. \\
\hline
\textbf{Out-of-Distribution (OOD) Detection} & Can the model trust its prediction on unseen data? & Uses confidence calibration, density estimation, and contrastive learning to flag OOD samples. & Agents actively request additional evidence for uncertain inputs and retrieve external knowledge dynamically to improve robustness. \\
\hline
\end{tabular}}
\end{table*}

\subsection{What questions and research directions do data-savvy agents unlock?}

The introduction of data-savvy agents expands the research landscape beyond what is possible with traditional Data-Centric AI approaches. We highlight the link to the main text below.

\textbf{Proactive Data Acquisition (Section 3)}: 

    \begin{itemize}
        \item How can AI systems autonomously discover and integrate new knowledge?
 \item Traditional AI systems rely on static pre-defined datasets, whereas Data-Savvy Agents can actively seek out missing data from dynamic, multi-source environments.
 \item As outlined in Section 3, this enables context-aware data acquisition, where agents move beyond passive querying and actively explore, retrieve, and structure data based on evolving needs.
    \end{itemize}

\newpage
\textbf{Sophisticated Data Processing (Section 4)}: 

  \begin{itemize}

 \item  How do AI systems process and refine messy, real-world data interactively?
\item Section 4 discusses the limitations of current methods, where AI agents tend to treat data processing as a fixed pipeline rather than an adaptive reasoning process.
\item Data-Savvy Agents diagnose interdependent data issues, reason about missingness, and actively correct errors through self-supervised feedback loops. This unlocks new research into adaptive data reasoning, allowing AI to refine its understanding rather than merely ingesting static datasets.

    \end{itemize}

 \textbf{Interactive Auto-Evaluation (Section 5)}: 

   \begin{itemize}

   \item Can AI systems generate their own evaluation data to test themselves?
\item Current ML evaluation relies on static benchmarks, making it difficult to measure AI systems in dynamic, evolving environments.
\item As outlined in Section 5, data-savvy agents introduce automated and context-aware test generation, where agents simulate interactions, generate counterfactuals, and iteratively refine their own evaluation. This enables interactive evaluation paradigms that go beyond pre-defined metrics, allowing AI to self-assess its generalization capabilities.

    \end{itemize}

\textbf{Continual Adaptation (Section 6)}: 

  \begin{itemize}

  \item How do AI systems remain up-to-date without full retraining?
\item Static AI models fail when data distributions shift, requiring expensive re-training.
\item As discussed in Section 6, current methods struggle with catastrophic forgetting and reactive rather than proactive adaptation.
Data-Savvy Agents introduce incremental knowledge updating, allowing AI to retain prior knowledge while seamlessly integrating new information. This shifts towards truly lifelong learning frameworks.
    \end{itemize}

Overall, Data-Savvy Agents transform AI’s relationship with data, shifting from passive dataset curation to active, autonomous data reasoning. Instead of relying on static training data, these agents continuously acquire, validate, and refine their knowledge in response to real-world conditions.

\subsection{Related Fields}
Beyond the literature discussed in the main body, we highlight (or call back to) several areas within the ML community that are related to \emph{certain aspects} of the capabilities of the proposed data-savvy agent.

\paragraph{Continual Learning.}
Continual learning, also known as lifelong learning, refers to the ability of a model to learn continuously from new data without forgetting previously acquired knowledge. This field has seen substantial progress in recent years, with techniques designed to address challenges such as catastrophic forgetting and the integration of new knowledge over time~\citep{van2019three,lee2020clinical, wang2024comprehensive}. 
However, while continual learning is a critical component of data-savvy agents, it cannot directly address the complex and dynamic demands of continual adaptation in data-savvy agents. 
The main gap lies in the difference between learning and adapting to new information. 
In continual learning, the focus is primarily on incremental knowledge updates within a static or predefined task space. 
The assumption is that the data distribution remains relatively stable and the agent’s tasks are well-defined.

For data-savvy agents, however, continual adaptation involves a more proactive and flexible approach. These agents need to not only learn from new data but also autonomously adjust their decision-making processes, knowledge structures, and interactions with external systems in response to real-time changes in the environment. This includes:
\begin{itemize}[nosep]
    \item[1.] Proactive Data Acquisition: Data-savvy agents must not only incorporate new data but also actively acquire data based on the current needs of the system, which may involve identifying gaps in knowledge or sensing when changes in the environment require adaptation.
	\item[2.] Dynamic Goal Adjustment: Unlike traditional continual learning, where the learning process follows a predefined objective, continual adaptation for data-savvy agents requires frequent realignment of goals based on shifting user needs and evolving task environments.
    \item[3.] Multimodal Integration: Data-savvy agents often work with diverse, real-time data sources (e.g., sensor data, user feedback, or interaction logs). Continual adaptation requires a seamless integration of this heterogeneous information into a unified model, something traditional continual learning methods often struggle with.
    \item[4.] Adaptation to Changing Environments: Data-savvy agents operate in non-stationary, ever-changing environments where the data distribution, task requirements, and even the problem definitions are subject to rapid shifts. Continual learning techniques are typically not designed to handle such dynamic and unpredictable changes in real time.
\end{itemize}
	
Thus, while continual learning lays the foundation for maintaining and updating knowledge over time, it does not inherently account for the proactive, flexible, and context-aware adaptations required by data-savvy agents in real-world applications. 
Addressing this gap necessitates novel approaches that extend beyond the current scope of continual learning to incorporate real-time, context-aware adaptation and decision-making.

\paragraph{Active Learning.}
Please refer to the ``Traditional Data Acquisition via Active Learning'' in Section~\ref{subsec:data-acquisition-limitations}.

\paragraph{Data-Centric Tools.}
Please refer to the ``Current Data-Centric Tools'' in Section~\ref{subsec:processing-limitations} as well as Table~\ref{table:data_comparison}.

\clearpage
\newpage

\section{More Examples on Real-World Impacts and Agent Design Blueprints}
\label{appendix:impact}
In addition to Section~\ref{sec:impact}, we put more examples on real-world impacts of data-savvy agents.

\subsection{Autonomous Policy Adaptation}

It is challenging for governments and global institutions to keep up with rapidly changing socio-economic and environmental situations. 
A data-savvy agent could transform policy-making by continuously analyzing real-time data, synthesizing insights, and generating adaptive policy recommendations.
Traditional policy-making relies on slow, periodic data collection and expert analysis, making it difficult to respond quickly to crises. 
A data-savvy agent could proactively gather policy-relevant data, simulate different decisions, and refine recommendations based on real-world feedback.

For example, in climate policy, an AI-driven system could analyze global emissions, predict the impact of carbon reduction strategies, and adjust regulations based on new scientific findings. In economic planning, it could detect financial instability early, recommend countermeasures, and fine-tune fiscal policies in real time. Here we provide the agent design blueprint for this application:
\paragraph{(1) Proactive Data Acquisition}
\begin{itemize}
    \item Scrape government sites, legal databases, real-time economic/social indicators.
    \item Interview policy analysts and integrate expert commentary.
\end{itemize}

\paragraph{(2) Sophisticated Data Processing}
\begin{itemize}
    \item Parse legal documents and identify causality (e.g., tax law $\rightarrow$ SME failure).
    \item Filter misinformation, outdated regulations, and biased media.
\end{itemize}

\paragraph{(3) Interactive Test Data Synthesis}
\begin{itemize}
    \item Generate policy simulation scenarios (``what-if'' cases).
    \item Create synthetic citizen feedback datasets to test impact projections.
\end{itemize}

\paragraph{(4) Continual Adaptation}
\begin{itemize}
    \item Adjust models with shifting public opinion, new laws, or regulatory updates.
    \item Track geopolitical signals and economic volatility for early warning.
\end{itemize}

\subsection{Personalized and Lifelong Education}
Education systems struggle to provide personalized learning at scale. A data-savvy agent could enable AI-driven lifelong education by adapting to individual learners and optimizing curricula.
Traditional education relies on standardized curricula and fixed assessments, which often fail to accommodate different learning paces and styles. 
A data-savvy agent could continuously acquire knowledge across disciplines, update teaching strategies based on cognitive science, and personalize learning through interactive auto-evaluation.
For example, in K-12 education, an AI tutor could adjust lessons in real time based on a child’s progress, offering personalized exercises and explanations. 
In addition, we provide the agent design blueprint for personalized and lifelong education as follows:
\paragraph{(1) Proactive Data Acquisition}
\begin{itemize}
    \item Collect learning logs from learning management system, assignment platforms, and interaction history.
    \item Query students/teachers for goals, gaps, and interests.
    \item Scrape domain-specific resources (Khan Academy, Coursera) and tag content by learning intent.
\end{itemize}

\paragraph{(2) Sophisticated Data Processing}
\begin{itemize}
    \item Align multimodal data (e.g., videos, tests, notes) and identify learning bottlenecks.
    \item Filter out noise (e.g., meaningless clicks, AI-generated fake reviews).
\end{itemize}

\paragraph{(3) Interactive Test Data Synthesis}
\begin{itemize}
    \item Generate custom quizzes based on current mastery level.
    \item Create interactive dialogues simulating Socratic-style questioning.
\end{itemize}

\paragraph{(4) Continual Adaptation}
\begin{itemize}
    \item Update teaching strategy based on engagement and feedback.
    \item Monitor curriculum alignment with new education standards.
\end{itemize}

At a global level, AI-driven education could democratize access to high-quality, evolving learning resources, ensuring personalized education for all, regardless of location or background. This shift could revolutionize workforce development, accelerate innovation, and bridge global knowledge gaps.

\subsection{Precision Healthcare}
Healthcare is a data-intensive field, yet inefficiencies persist in diagnosis, treatment, and research.
A data-savvy agent could revolutionize precision medicine, clinical decision-making, and drug discovery by continuously acquiring, processing, and evaluating medical data at an unprecedented scale.
Traditional medical research relies on manual data collection, time-consuming clinical trials, and retrospective analysis, often leading to slow innovation cycles. 
A data-savvy agent could proactively acquire patient data from diverse sources (genomic data, wearable devices, EHRs, clinical studies) and generate real-time insights. 
Through interactive auto-evaluation, it could refine disease models, simulate drug interactions, and optimize treatment protocols based on real-world patient outcomes.

For example, in oncology, an AI-driven system could dynamically personalize cancer treatment plans by integrating real-time patient responses with the latest clinical research. In drug development, it could simulate biochemical interactions, drastically reducing the time and cost required to bring new therapies to market. Here we provide the agent design blueprint for this application:
\paragraph{(1) Proactive Data Acquisition}
\begin{itemize}
    \item Integrate EMRs, wearables, lab results, and clinician notes.
    \item Crawl up-to-date drug guidelines and clinical trial data.
\end{itemize}

\paragraph{(2) Sophisticated Data Processing}
\begin{itemize}
    \item Merge structured and unstructured medical data (PDFs, scans, lab tables).
    \item Flag biases (e.g., underrepresented demographics) and missing features.
\end{itemize}

\paragraph{(3) Interactive Test Data Synthesis}
\begin{itemize}
    \item Generate synthetic patient profiles for diagnosis stress tests.
    \item Create rare-case test sets to probe model edge performance.
\end{itemize}

\paragraph{(4) Continual Adaptation}
\begin{itemize}
    \item Adjust recommendations in response to new treatments or disease outbreaks.
    \item Track policy and regulation changes (HIPAA, insurance codes).
\end{itemize}

\subsection{Resilient Global Supply Chains}
In a volatile world, supply chains must rapidly adapt to disruptions from pandemics, geopolitics, and natural disasters. Traditional models, reliant on static data, struggle to anticipate sudden shocks. A data-savvy agent could continuously monitor trade flows, environmental conditions, and geopolitical shifts, dynamically adjusting strategies. With interactive auto-evaluation and continual adaptation, it could auto-simulate scenarios, test contingency plans, and enhance overall resilience, ensuring more responsive and robust global supply networks. For example, in global food supply chains, such a data-savvy agent could predict drought-related shortages and automatically adjust distribution to prevent famine. Here we provide the agent design blueprint for this application:
\paragraph{(1) Proactive Data Acquisition}
\begin{itemize}
    \item Fetch real-time logistics data from sensors, customs reports, satellite imagery.
    \item Extract procurement status from internal ERP systems.
\end{itemize}

\paragraph{(2) Sophisticated Data Processing}
\begin{itemize}
    \item Align disparate timescales (hourly shipping vs monthly forecasts).
    \item Detect root causes of delay (e.g., labor strike vs policy vs storm).
\end{itemize}

\paragraph{(3) Interactive Test Data Synthesis}
\begin{itemize}
    \item Simulate events like factory outage, port closure, or demand spikes.
    \item Use digital twins to test AI-based scheduling systems.
\end{itemize}

\paragraph{(4) Continual Adaptation}
\begin{itemize}
    \item React to seasonality, currency shifts, or trade policy changes.
    \item Update sourcing and risk assessment strategy continuously.
\end{itemize}

\end{document}